\documentclass[10pt,journal,compsoc]{IEEEtran}

\ifCLASSOPTIONcompsoc
  \usepackage[nocompress]{cite}
\else
  \usepackage{cite}
\fi

\usepackage{amsmath,amsfonts}
\usepackage{algorithmic}
\usepackage{array}
\usepackage[caption=false,font=normalsize,labelfont=sf,textfont=sf]{subfig}
\usepackage{textcomp}
\usepackage{stfloats}
\usepackage{url}
\usepackage{verbatim}
\usepackage{graphicx}
\usepackage{cite}
\usepackage{bm}
\usepackage{balance}
\usepackage{color}
\usepackage{makecell}
\def\BibTeX{{\rm B\kern-.05em{\sc i\kern-.025em b}\kern-.08em
    T\kern-.1667em\lower.7ex\hbox{E}\kern-.125emX}}
\usepackage{balance}
\begin{document}
\title{DPFormer: Dynamic Prompt Transformer for Continual Learning}
\author{
Sheng-Kai Huang, Jiun-Feng Chang, and Chun-Rong Huang, \IEEEmembership{Senior Member, IEEE}

\IEEEcompsocitemizethanks{\IEEEcompsocthanksitem
Sheng-Kai Huang is with the Department of Software-GPU, Nvidia Corporation, Taipei 114, Taiwan. E-mail: shengkaih@nvidia.com
\IEEEcompsocthanksitem
Jiun-Feng Chang is with the Department of Computer Science and Engineering, National Chung Hsing University, Taichung 402, Taiwan. E-mail: g112056009@mail.nchu.edu.tw.
\IEEEcompsocthanksitem
Chun-Rong Huang is with the Department of Computer Science, National Yang Ming Chiao Tung University, Hsinchu 300, Taiwan, and the Department of Computer Science and Engineering, National Chung Hsing University, Taichung 402, Taiwan. E-mail: crhuang@cs.nycu.edu.tw.
}
\thanks{Manuscript received May 25, 2024. This work was supported in part by the National Science and Technology Council of Taiwan under Grants NSTC 112-2634-F-006-002 and NSTC 111-2628-E-A49-028-MY3.
\emph{(Corresponding author: Chun-Rong Huang.)}}
}

\markboth{Submitted to IEEE Transactions on Pattern Analysis and Machine Intelligence, 2024}%
{DPFormer: Dynamic Prompt Transformer for Continual Learning}

\IEEEtitleabstractindextext{
\begin{abstract}
%
%
In continual learning, solving the catastrophic forgetting problem may make the models fall into the stability-plasticity dilemma.
Moreover, inter-task confusion will also occur due to the lack of knowledge exchanges between different tasks.
In order to solve the aforementioned problems, we propose a novel dynamic prompt transformer (DPFormer) with prompt schemes.
The prompt schemes help the DPFormer memorize learned knowledge of previous classes and tasks, and keep on learning new knowledge from new classes and tasks under a single network structure with a nearly fixed number of model parameters.
%
%
%
Moreover, they also provide discrepant information to represent different tasks to solve the inter-task confusion problem. 
Based on prompt schemes, a unified classification module with the binary cross entropy loss, the knowledge distillation loss and the auxiliary loss is proposed to train the whole model in an end-to-end trainable manner. 
Compared with state-of-the-art methods, our method achieves the best performance in the CIFAR-100, ImageNet100 and ImageNet1K datasets under different class-incremental settings in continual learning.
The source code will be available at our GitHub after acceptance. 
\end{abstract}

}

\maketitle

\IEEEdisplaynontitleabstractindextext

%
\IEEEpeerreviewmaketitle

\section{Introduction}
\IEEEPARstart
{T}{he} purpose of continual learning is to enable the artificial intelligence model to learn new samples of new tasks without forgetting previously learned knowledge.
When models only consider the adaption to new samples, their performance on previous samples often significantly degrades.
This phenomenon is called catastrophic forgetting which is a major challenge in continual learning.
In order to avoid catastrophic forgetting, the models need to retain previously learned knowledge which may limit the model adaptability to learn new samples.
Therefore, adapting to new samples and retaining previously learned knowledge constitute the stability-plasticity dilemma~\cite{Grossberg_SMB_1982, Mermillod_FP_2013 ,Parisi_NN_2019} in the class-incremental learning (class-IL)~\cite{Masana_TPAMI_2023} problem, where the learners must discriminate all classes in the previous tasks during inference.

To achieve continual learning without catastrophic forgetting, traditional rehearsal methods~\cite{Rebuffi_CVPR_2017, Castro_ECCV_2018, Chaudhry_2018_ECCV, Rolnick_NIPS_2019, Prabhu_2020_ECCV,Boschini_TPAMI_2023} store previous samples in additional memory buffers.
When the model learns new samples, it will also retrain previous samples to prevent catastrophic forgetting.
When the number of tasks increases, the memory usage and retraining time of previous samples will also increase. 
%
To reduce the memory usage and retraining time, regularization methods~\cite{Kirkpatrick_PNAS_2017, Liu_ICPR_2018, Lee_CVPR_2020, Zenke_ICML_2017, Aljundi_2018_ECCV, Li_2018_TPAMI, Hou_2019_CVPR, Zhang_2020_WACV, Hou_CVPR_2019, Wu_CVPR_2019, Zhao_CVPR_2020, Douillard_ECCV_2020} are proposed.
These methods aim to control the update schemes of the models so that the models will not be updated by only considering new incoming samples of new tasks.
However, when more new tasks need to be learned, the update schemes will limit the models to focus on the previously learned knowledge and make the models hard to learn new knowledge.
Although regularization methods may prevent catastrophic forgetting, they generally fall into the stability-plasticity dilemma.
To address the stability-plasticity dilemma, dynamic architecture methods~\cite{Serra_ICML_2018, Ke_2021_NIPS, Masana_2021_CVPR, Zhang_2022_TNNLS, Verma_2021_CVPR, Hu_2021_AAAI, Rajasegaran_NIPS_2019, Yan_CVPR_2021, Li_arXiv_2021, Douillard_CVPR_2022} are proposed by adding additional networks to learn each new task.
In this way, new networks can learn new knowledge for new tasks while previous networks retain previously learned knowledge to effectively prevent catastrophic forgetting.
However, the number of parameters and calculations of the models will significantly increase with respect 
to the number of tasks.
Moreover, because all of the classes of the tasks are never jointly trained in class-IL, the networks are hard to simultaneously discriminate all classes and thus the inter-task confusion~\cite{Masana_TPAMI_2023} problem occurs.

To solve the aforementioned problems, we propose the dynamic prompt transformer (DPFormer).
Two unique prompt modules are proposed in the DPFormer to memorize the information of previous classes and tasks to avoid catastrophic forgetting.
When learning new tasks, these modules also simultaneously generate representative features for new classes and tasks to enhance the plasticity of the model. 
With the prompt modules, a single network structure with a nearly fixed number of model parameters can be achieved for solving the class-IL problem.  

%
%
The DPFormer contains a feature encoder module, a class prompt module, a task prompt module and a unified classification module.
The feature encoder module aims to extract spatially significant features to effectively represent each image.
The class prompt module learns new class prototypes for new classes in the new task to increase the plasticity of the model. 
Moreover, it selects the most representative class prototypes as the prompts for the input image to remind the network to avoid the catastrophic forgetting for each class. 
To represent the discrepancy between classes of different tasks in class-IL, the task prompt module is proposed.
It learns new task prototypes for each task and selects the most representative task prototypes as the prompts for the input image to describe the discrepancy between tasks. 
With these two modules, our method can not only solve the catastrophic forgetting problem but also overcome the stability-plasticity dilemma.

Conventional dynamic architecture methods usually require multiple classifiers for different tasks, which leads to the lack of knowledge exchange among classifiers and causes the inter-task confusion problem.
To solve the problem, we only use one unified classification module with a standard classifier and an auxiliary classifier based on selected representative class and task prototypes of the class and task prompt modules. 
To train the network in an end-to-end trainable manner, a binary cross entropy loss, a knowledge distillation loss, and an auxiliary loss are proposed. 
The binary cross entropy loss aims to learn to classify all samples, while the knowledge distillation loss helps increase the stability of the model to avoid catastrophic forgetting.
To further improve the plasticity of the model of the current task, the auxiliary loss is computed based on the auxiliary classifier.
Our unified classification module takes into account the knowledge between different tasks and makes our model more capable of distinguishing classes in different tasks.
By considering prompt modules and a unified classification module, our method is more suitable to solve the class-IL problem with a large number of tasks.

In the preliminary version~\cite{Huang_MVA_2023}, a concept of task selection module (TSM) is proposed to achieve class-IL with a single classifier.
Selecting proper task token is hard to discriminate samples of new and previous classes, and thus TSM~\cite{Huang_MVA_2023} still suffers from catastrophic forgetting.
Compared with~\cite{Huang_MVA_2023}, our method further considers neighborhood attention based feature extraction to extract more reliable patch features, the class prompt module to prompt class information of different tasks, the task prompt module to describe inter-task discrepancy, and the auxiliary classifier for learning new classes to increase model plasticity. 
As a result, the proposed method is significantly superior to~\cite{Huang_MVA_2023}.


The contribution of this paper is four-fold. 
\begin{itemize}
    \item We propose the dynamic prompt transformer (DPFormer) which can effectively retain previously learned knowledge by using the class prompt module and task prompt module to avoid catastrophic forgetting and learn new knowledge of new tasks to improve the plasticity of the model. 
    Therefore, our method can overcome the stability-plasticity dilemma in continual learning.
    \item The class prompt module and task prompt module are cooperated to provide informative prompts for each image to boost the classification performance in continual learning.
    \item A unified classification module is proposed to classify images of different tasks which avoid the significant parameter increment with respect to the number of tasks.
    Moreover, it can also take into account the knowledge from all tasks to avoid inter-task confusion. 
    \item The proposed method outperforms the state-of-the-art rehearsal methods, regularization methods, and dynamic architecture methods in the CIFAR-100, ImageNet100 and ImageNet1K datasets under different class-IL settings in continual learning. 
\end{itemize}

The paper is organized as follows. 
Sec. II gives the related work of continual learning methods. 
The proposed method is presented in Sec. III. 
Sec. IV shows the experimental results and comparisons. 
Finally, the conclusions are given in Sec. V.

\section{Related Work}
In this section, we will review prior works in continual learning including rehearsal methods, regularization methods and dynamic architecture methods. 

\subsection{Rehearsal Methods}
To prevent catastrophic forgetting, a naive idea is to use additional memory buffers to store samples of previous classes. 
When learning a new task, the stored samples in the buffers can be retrained to prevent the model from forgetting previous knowledge.
Rebuffi et al.~\cite{Rebuffi_CVPR_2017} propose an incremental classifier and representation learning (iCaRL) method by using a nearest-mean-of-exemplars to select representative samples of each seen class and a fixed memory buffer to store the old exemplars. 
During training, they regularly update representative samples and retrain the old exemplars to prevent catastrophic forgetting.
Castro et al.~\cite{Castro_ECCV_2018} propose an end-to-end incremental learning (E2E) by using herding selection~\cite{Welling_ICML_2009} strategy to select most representative samples based on the distance to the mean sample of each class.
In~\cite{Chaudhry_2018_ECCV}, they introduce metrics to quantify catastrophic forgetting and intransigence by selecting exemplars with higher entropy, and selecting exemplars based on how close they are to the decision boundary, respectively.
Rolnick et al.~\cite{Rolnick_NIPS_2019} propose continual learning with experience and replay (CLEAR) by mixing on-policy learning for plasticity and off-policy learning from replay experiences to maintain stability.
Prabhu et al.~\cite{Prabhu_2020_ECCV} assume a balanced training set, greedily store samples in memory in a class-balanced way, and train models from scratch by using only samples in the memory. 
Boschini et al.~\cite{Boschini_TPAMI_2023} propose extended dark experience replay to maintain the memory buffer up-to-date by considering current 
information into the memories of the previous samples and  evaluate the benefits of preparing the underlying model to incoming tasks.

\subsection{Regularization Methods}
Due to the limited size of the memory buffers, the performance of the rehearsal methods is limited. 
Moreover, the larger buffer size implies more training time.
Therefore, regularization methods are proposed to reserve previously learned knowledge by limiting the update of the models.
For example, Kirkpatrick et al.~\cite{Kirkpatrick_PNAS_2017} propose elastic weight consolidation (EWC) which penalizes important parameters if they are updated during training for new tasks to limit the update direction of the model and prevent the model from being too inclined to new data.
Liu et al.~\cite{Liu_ICPR_2018} improve EWC by rotating the parameter space to provide a better approximation of the Fisher information matrix.
However, the model has to be extended with more parameters during training, which does not increase the capacity of the network but incurs in a computational and memory cost.
Similar to~\cite{Liu_ICPR_2018}, Lee et al.~\cite{Lee_CVPR_2020} propose an extension of the Kronecker factorization technique for block-diagonal approximation of the Fisher information matrix to handle the model update. 
%
Zenke et al.~\cite{Zenke_ICML_2017} propose the path integral approach (PathInt), that accumulates the changes in each parameter online along the entire learning trajectory.
As shown in~\cite{Zenke_ICML_2017}, batch updates to the weights might lead to overestimating the importance, while starting from pre-trained models might lead to underestimating it.
To address this issue, memory aware synapses (MAS)~\cite{Aljundi_2018_ECCV} is proposed to calculate importance values for each network weight online by accumulating the sensitivity of the learning function, i.e. the magnitude of the gradient of the learning function.

Some regularization-based methods aim to prevent activation drift based on knowledge distillation which was originally designed to learn a more compact student network from a larger teacher network.
Li et al.~\cite{Li_2018_TPAMI} propose to use knowledge distillation to keep the representations of samples of previous classes from drifting too much while learning new tasks.
Instead of regularizing network predictions, Hou et al.~\cite{Hou_2019_CVPR} propose the less-forget constraint
to regularize the cosine similarity between the L2-normalized logits of the previous and current network.
Zhang et al.~\cite{Zhang_2020_WACV} propose the deep model consolidation (DMC) based on the observation that there exists an asymmetry between previous and new classes when training.
New classes have explicit and strong supervision, whereas supervision for previous classes is weaker and communicated by means of knowledge distillation.

More recently, Hou et al.~\cite{Hou_CVPR_2019} propose learning a unified classifier incrementally via rebalancing (UCIR) by using cosine normalization to balance magnitudes across all classes and design less-forget constraint to preserve the geometric configuration of old classes with the distillation loss and the margin ranking loss.
Wu et al.~\cite{Wu_CVPR_2019} propose bias correction (BiC) by introducing two stage training. 
In the first stage, the new task is trained with the cross-entropy loss and the knowledge distillation loss, while in the second stage, they freeze all the parameters in the network and add a bias correction layer after the last fully connected layer to avoid the model to be biased by new classes.
%
Zhao et al.~\cite{Zhao_CVPR_2020} propose the weight aligning (WA) to maintain the fairness of old and new classes by correcting the biased weights in the fully connected layer during training. 
Douillard et al.~\cite{Douillard_ECCV_2020} propose pooled outputs distillation network (PODNet) to balance the compromise between preserving previous knowledge and learning new classes by using an efficient spatial-based distillation loss to remain the stability of network.

\subsection{Dynamic Architecture Methods}
Regularization approaches aim to prevent catastrophic forgetting by limiting the update of the models which may sacrifice the model adaptability for learning new data and fall into the stability-plasticity dilemma. 
To address the stability-plasticity dilemma, dynamic architecture methods are proposed.
For example, Serra et al.~\cite{Serra_ICML_2018} propose the hard attention to the task (HAT) mechanism to preserve the previous task knowledge by learning additional attention vectors for each task.
Rajasegaran et al.~\cite{Rajasegaran_NIPS_2019} propose random path selection network (RPSNet).
For each new task, a set of candidate path is randomly sampled, and parallel encoders are required to generate candidate paths.
Yan et al.~\cite{Yan_CVPR_2021} propose a two-stage strategy including dynamically expandable representation (DER) learning and classifier learning.
To prevent catastrophic forgetting, they fix the previous feature representation and dynamically add new feature extractors for new tasks to learn the feature representations to increase model plasticity.
However, the numbers of model parameters of these methods will significantly increase with respect to the number of tasks.

To overcome the number of parameters problem, Li et al.~\cite{Li_arXiv_2021} propose using knowledge distillation and a simpler pruning procedure to DER~\cite{Yan_CVPR_2021}. 
Verma et al.~\cite{Verma_2021_CVPR} propose task-specific feature map transformation to learn a transformation for each new task by adding a regularizer to force the new task features to be separated from those learned in the previous tasks.
%
%
Per-class continual learning (PCL)~\cite{Hu_2021_AAAI} dynamically adds separate heads of the model for each class in new task, and they apply one class loss to learn one class.
Because each head only recognizes the target class, the feature representations of different classes may be similar and lead to the inter-task confusion problem.
Zhang et al.~\cite{Zhang_2022_TNNLS} propose a self-growing binary activation network (SGBAN), which can progressively extend the neurons in the fully connected network to increase model plasticity, but the performance of this method is limited by the baseline architecture.
Douillard et al.~\cite{Douillard_CVPR_2022} leverage the vision transformer~\cite{Dosovitskiy_ICLR_2021} to propose dynamic token expansion (DyTox) which  learns new tasks via the dynamic task decoder.
For each task, they add new task-specific learned tokens to learn the information in each task and create new classifiers to classify samples in each task.
Nevertheless, separating task decoders and using multiple classifiers will lead to inter-task confusion and thus reduce the classification performance.
For more reviews, please refer to~\cite{Lange_TPAMI_2022} and ~\cite{Wang_TPAMI_2024}.

To address aforementioned problems, we propose the DPFormer to prevent catastrophic forgetting via class and task prototypes.
Previously learned knowledge can be preserved in class and task prototypes which enhance the stability of the model.
On the other hand, the newly added prototypes can be used to learn new knowledge and improve the plasticity of the model.
Then, our unified classification module can simultaneously discriminate all classes by considering the learned prototypes and patch tokens to avoid inter-task confusion.
In addition, dynamically learning prototypes for classes and tasks only requires a small amount of parameters which makes the model be trained with a nearly fixed number of model parameters.
Therefore, DPFormer can not only overcome the stability-plasticity dilemma, but also be applied to a large number of tasks.
\begin{figure*}[!t]
\centering
\includegraphics[width=\textwidth]{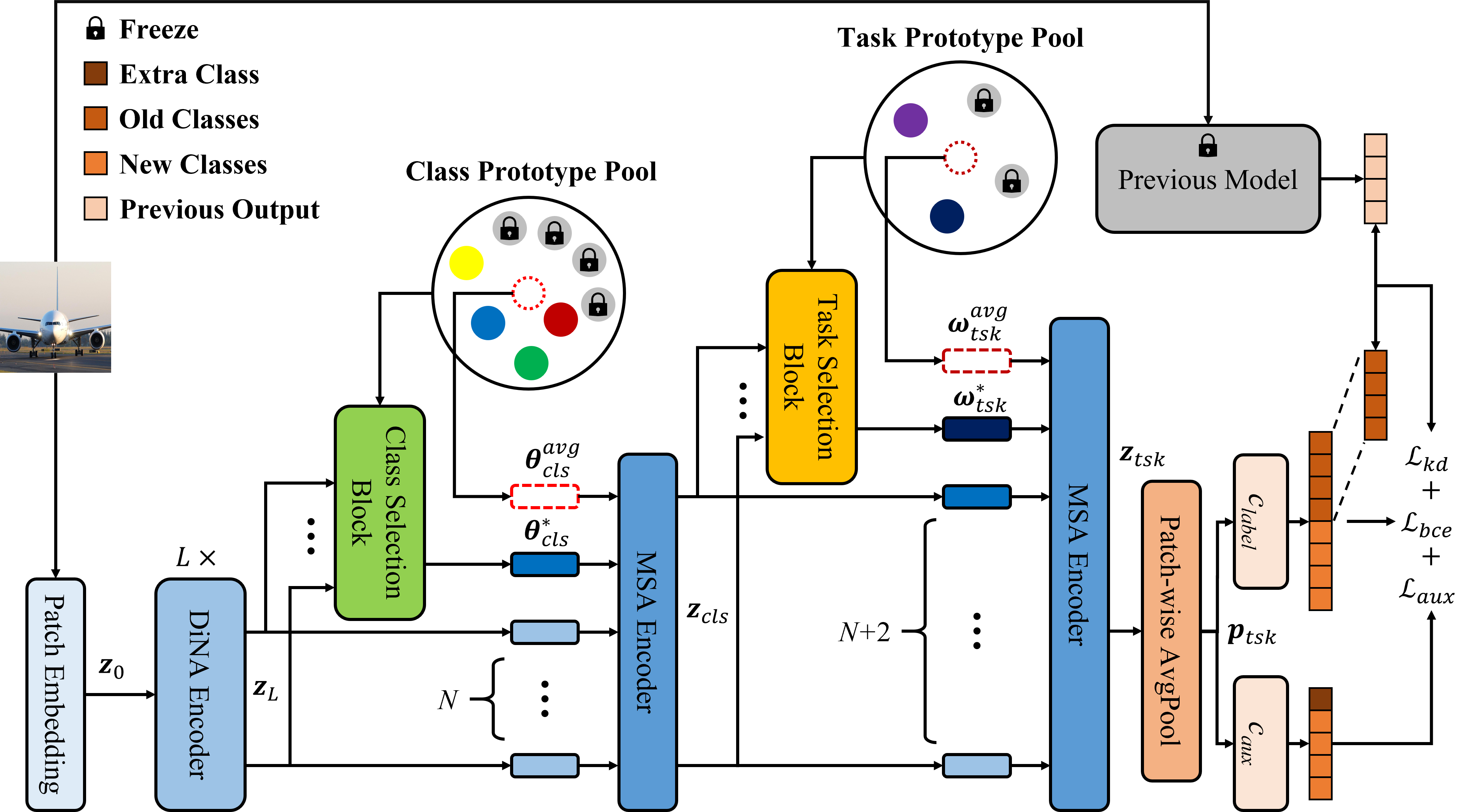}
\caption{The overview of the DPFormer. 
The DPFormer consists of four modules including the feature encoder module, the class prompt module, the task prompt module and the unified classification module.
The feature encoder module aims to learn transformer features $\bm{z}_L$ based on the dilated neighborhood attention (DiNA) scheme from patch tokens $\bm{z}_0$ of an input image $\bm{x}^t_i$ of task $t$. 
The class prompt module selects the most representative class prototype to remind the model about previously learned knowledge to avoid catastrophic forgetting.
The task prompt module aims to represent
the discrepancy between different tasks by learning task knowledge. 
The unified classification module contains two classifiers, the label classifier $c_{label}$ and the auxiliary classifier $c_{aux}$ to compute the binary cross entropy loss $\mathcal{L}_{bce}$, the knowledge distillation loss $\mathcal{L}_{kd}$, and the auxiliary loss $\mathcal{L}_{aux}$ for training the model. 
}
\label{fig:flowchart}
\end{figure*}

\section{Method}
%
In class-IL setting of continual learning, given a task $t$ from a task set $\mathcal{T} = \{1, ..., T\}$, it contains $N^t$ training samples $\{\bm{x}^t_i, y^t_i\}^{N^t}_{i=1}$, where $\bm{x}^t_i$ is the $i$th training sample in task $t$, $y^t_i$ is the label of $\bm{x}^t_i$ and $y^t_i$ belongs to the class set $C^t$ in task $t$. 
Given any two different tasks $t$ and $t'$, the class sets $C^{t}$ and $C^{t'}$ of tasks $t$ and $t'$ are disjoint, i.e. $C^{t} \cap C^{t'} = \{\emptyset | \forall t \neq t'$\}.
%
%
The goal of the paper is to train a model to classify new incoming samples and maintain the ability to classify all learned classes, i.e., classify all samples of classes $C^{1:t}$ from the testing set.

\subsection{Overview}
The proposed framework is shown in Fig.~\ref{fig:flowchart}. 
%
Our network consists of four modules including the feature encoder module, the class prompt module, the task prompt module and the unified classification module.
The feature encoder module aims to learn transformer features from the patch tokens $\bm{z}_0$ of an input image $\bm{x}^t_i$ of task $t$. 
%
%
Based on the learned features, the class prompt module will select proper class prototypes which represent the learned class knowledge and concatenate these prototypes with the patch tokens of the input image.
A transformer encoder is then applied to discover the correlations between class prototypes and patch tokens.
The class prompt module further learns representative class prototypes for each new class from training samples in new task $t$. 
The task prompt module aims to represent
the discrepancy between different tasks.
It will select a proper task prototype by further considering the class prototype because each task prototype contains the class information learned in the task.
The selected task prototype representing discrepant task knowledge for $\bm{x}^t_i$ is concatenated with the features of the class prompt module to discover the correlations between patch tokens, class prototypes, and task prototypes by using a transformer encoder.
After the task prompt module, to extract the global representation of $\bm{x}^t_i$, we perform the patch-wise average pooling to obtain the dominant feature $\bm{p}_{tsk}$, which serves as the input of the unified classification module.  

The unified classification module contains two classifiers, the label classifier and the auxiliary classifier. 
The label classifier $c_{label}$ is used to predict the class probability $\hat{y}^{1:t}_i$ of $\bm{x}^t_i$. 
%
The auxiliary classifier $c_{aux}$ is used to distinguish if $\bm{x}^t_i$ is from the classes of the previous tasks or belongs to the classes of the current task.
To drive the learning of the model, the loss function $\mathcal{L}$ is defined as follows:
\begin{equation}
   \mathcal{L} = (1-\alpha)\mathcal{L}_{bce} + \alpha\mathcal{L}_{kd} + \lambda\mathcal{L}_{aux},
\end{equation}
where $\alpha = \frac{|C^{1:t-1}|}{|C^{1:t}|}$ is the fraction of the total number of classes in task $t-1$ over the total number of classes in task $t$ and $\lambda$ is a hyperparameter.
%
%
%
The binary cross entropy loss $\mathcal{L}_{bce}$ is computed based on the output of $c_{label}$ and ground truth. 
To remember the knowledge of the previous tasks, the learned model in task $t-1$ is also used to predict the probability of the samples of task $t$ during training.
The knowledge distillation loss $\mathcal{L}_{kd}$~\cite{Geoffrey_Hinton_NIPSW_2015} is computed based on the outputs of the models of task $t$ and $t-1$. 
Final, the auxiliary loss $\mathcal{L}_{aux}$ is computed based on the output of $c_{aux}$ to learn the diversity among previous and current tasks~\cite{Douillard_CVPR_2022}.
During inference, the label of the testing sample is obtained only based on the output of $c_{label}$.
In the following, we will introduce each component in the proposed method.

\subsection{Dynamic Prompt Transformer}

\subsubsection{Feature Encoder Module}
%
As shown in~\cite{Touvron_2021_ICML}, when the feature encoder uses only multi-head self-attention (MSA)~\cite{Dosovitskiy_ICLR_2021}, it is easy to overfit. 
%
To prevent overfitting, Swin transformer~\cite{Liu_2021_ICCV} considers a hierarchical structure with the shifted windows attention mechanism.
To further reduce the overfitting, the MSA scheme is replaced by the nearest neighborhood attention (NA)~\cite{Hassani_2023_CVPR}. 
NA is cooperated with dilation schemes (DiNA) to help fine-to-coarse feature learning~\cite{Hassani_arXiv_2022}.
Therefore, we apply the DiNA in our transformer encoder to obtain representative features.

Given an input image, we first pass the image through an overlapping tokenizer to obtain a patch embedding sequence $\bm{z}_0 \in \mathbb{R}^{N\times D}$, where $N$ represents the number of patches and $D$ represents the embedding dimension.
Then, the patch embedding sequence $\bm{z}_0$ serves as the input to the following transformer encoder.
The encoder network contains a total of $L$ dilated neighborhood transformer encoders to generate deep features.
Each dilated neighborhood transformer encoder consists of a multi-head dilated neighborhood attention block $DiNA(\cdot)$ and a multi-layer perceptron block $MLP(\cdot)$.
For the $l$th dilated neighborhood transformer encoder, the input $\bm{z}_{l-1}$ is passed to a normalization layer $LN(\cdot)$.
The normalized features are passed to the DiNA block with a residual link as follows: 
\begin{equation}
\bm{\hat{z}}_{l-1} = DiNA(LN(\bm{z}_{l-1}))+\bm{z}_{l-1}. 
\label{equ:dina1}
\end{equation}
%
%
The output of the DiNA block is passed to a normalization layer and the MLP block with a residual link as follows:
\begin{equation}
\bm{{z}}_l = MLP(LN(\bm{\hat{z}}_{l-1}))+\bm{\hat{z}}_{l-1}, 
\label{equ:dina2}
\end{equation}
where $\bm{z}_l$ is the output of the $l$th dilated neighborhood attention transformer encoder. 
Finally, $\bm{z}_L$ serves as the input of the class prompt module.

\subsubsection{Class Prompt Module}
To overcome the catastrophic forgetting problem and the stability-plasticity dilemma, we propose the class prompt module which provides learned class information to assist the representation of the input image. 
The class prompt module contains a class prototype selection block, a class prototype pool and a transformer encoder. 
The class prototype selection block aims to select the most representative class prototype from the class prototype pool based on the patch tokens generated by the encoder module. 
The most representative class prototype serves as the prompt for the input image to remind the model to avoid catastrophic forgetting. 

The class prototype pool contains class prototypes of two different sets. 
The first set is the previous class prototype set  
$\mathcal{\theta}^{pre}_{cls} = \{ \bm{\theta}_1, ..., \bm{\theta_}{C_{N}}\}$, where $C_N$ is the number of the previous class prototypes. 
$\mathcal{\theta}^{pre}_{cls}$ contains the class prototypes learned from task $1$ to task $t-1$. 
The second set is the current class prototype set  
$\mathcal{\theta}^{cur}_{cls} = \{ 
\hat{\bm{\theta}}_1, ..., \hat{\bm{\theta}}_{C_{M}}\}$, where $C_M$ is the number of the current class prototypes. 
$\mathcal{\theta}^{cur}_{cls}$ contains the class prototypes that need to be learned in task $t$. 
The previous class prototypes aim to preserve learned knowledge of previous classes from task $1$ to task $t-1$ for stability. 
On the other hand, the current class prototypes aim to learn new knowledge from current classes of task $t$ for plasticity. 

Given $\bm{z}_L$ generated by the feature encoder module, we perform the patch-wise average pooling to obtain the pooling feature $\bm{p}_L$ as follows:
\begin{equation}
    \bm{p}_L = \frac{1}{N} \sum_{n=1}^N \bm{z}^n_L,
\end{equation}
where $\bm{z}^n_L$ is the $n$th patch token of $\bm{z}_L$.
In this way, we can obtain the representative feature of the input image. 
The class prototype selection block will compare $\bm{p}_L$ with the class prototypes in the class prototype pool to find the most representative  class prototype as follows:
\begin{equation}
    \bm{\theta}^*_{cls}  = \mathop{\arg \max}\limits_{\bm{\theta} \in \{\mathcal{\theta}^{pre}_{cls} \bigcup \mathcal{\theta}^{cur}_{cls}\}} cs(\bm{p}_L, \bm{\theta}),
\end{equation}
where $\bm{\theta}^*_{cls}$ represents the selected class prototype and $cs(\cdot)$ represents the cosine similarity function. 
$\bm{\theta}^*_{cls}$ provides the most representative class information to remind the network of the correlations between the input image and the learned class knowledge to avoid catastrophic forgetting. 
%
Besides $\bm{\theta}^*_{cls}$, we also use the global information stored in the class prototype pool to maintain the stability of the model. 
Thus, the global information is represented by using the average class prototype $\bm{\theta}^{avg}_{cls}$ of the class prototypes in the class prototype pool and is defined as follows:
\begin{equation}
    \bm{\theta}^{avg}_{cls} = \frac{1}{C_N+C_M} (\sum_{j=1}^{C_N}  \bm{\theta}_j + \sum_{k=1}^{C_M} \hat{\bm{\theta}}_k),
\end{equation}
where the dimension of $\bm{\theta}^{avg}_{cls}$ is $\mathbb{R}^{1 \times D}$, $\bm{\theta}_j \in \mathcal{\theta}^{pre}_{cls}$, and  $\hat{\bm{\theta}}_k \in \mathcal{\theta}^{cur}_{cls}$.
The selected class prototype $\bm{\theta}^*_{cls}$ and the average class prototype $\bm{\theta}^{avg}_{cls}$ are concatenated to the patch tokens $\bm{z}_L$ to generate $\bm{z'}_{cls}$ as follows:
\begin{equation}
    \bm{z'}_{cls} = \bm{\theta}^*_{cls} \oplus \bm{\theta}^{avg}_{cls} \oplus \bm{z}_L,
\end{equation}
where $\bm{z'}_{cls} \in \mathbb{R}^{(N+2) \times D}$ and $\oplus$ is the concatenation operator.

%
%
Because the prototypes provide additional class information, we use the MSA~\cite{Dosovitskiy_ICLR_2021} scheme instead of DiNA~\cite{Hassani_arXiv_2022} in the class prompt module to discover the correlations of these prototypes with respect to all of the patch tokens.
%
The class prompt module contains a transformer encoder with a MSA block and a MLP block. 
Given $\bm{z'}_{cls}$, the learned features after the MSA block in the transformer encoder is denoted as follows:
\begin{equation}
\bm{\hat{z}}_{cls} = MSA(LN(\bm{z'}_{cls}))+\bm{z'}_{cls}.
\end{equation}
$\bm{\hat{z}}_{cls}$ is then passed to the MLP block to obtain the learned feature $\bm{z}_{cls}$ with class prototype information as follows:
\begin{equation}
\bm{z}_{cls} = MLP(LN(\bm{\hat{z}}_{cls}))+\bm{\hat{z}}_{cls}.
\end{equation}
%


%
In the first task, the class prototype pool has class prototypes with quantity $|C^1|$ generated by using the Gaussian distribution and these class prototypes will be updated, where the dimension of each class prototype is $\mathbb{R}^{1 \times D}$. 
In task $t$, we create new class prototype set $\theta^{cur}_{cls}$ with quantity $|C_M|$ and the previous class prototype set $\theta^{pre}_{cls}$ with quantity $|C_N|$ is frozen. 
When the transformer encoder of the class prompt module is updated, class prototypes in the new class prototype set will be updated to learn the class information of task $t$. 
In this way, the class prompt module can help remind the network the previously learned class knowledge and keep on learning new knowledge from new classes of new tasks. 
As a result, it can solve the catastrophic forgetting problem and overcome the stability-plasticity dilemma.

\subsubsection{Task Prompt Module}
To prevent inter-task confusion, the task prompt module is proposed which learns new task prototypes for each task and selects the most representative task prototype as the prompt for the input image to describe the discrepancy between tasks. 
The task prompt module contains a task selection block, a task prototype pool, and a transformer encoder.
The task selection block aims to select a representative task prototype for each input image based on the output of the class prompt module and patch tokens of the input image to prompt the task knowledge for the following unified classification module.
The task prototype pool stores the task prototypes and each task prototype represents the task information learned from images of classes of each task during training. 
In this way, we can use only a unified classification module to achieve continual learning. 
Finally, a transformer encoder is used to learn the features which represent the correlations among selected task prototypes, selected class prototypes and patch tokens for classification. 

%

%
The task prototype pool contains a previous task prototype set $\mathcal{\omega}^{pre}_{tsk} = \{ \bm{\omega}_1, ..., \bm{\omega}_{t-1}\}$ and the current task prototype $\bm{\omega}_{t}$ and the dimension of each task prototype is $\mathbb{R}^{1 \times D}$. 
%
%
%
The task selection block first performs patch-wise average pooling to obtain the pooling feature $\bm{p}_{cls}$ based on $\bm{z}_{cls}$ as follows:  
\begin{equation}
    \bm{p}_{cls} = \frac{1}{N+2} \sum_{n=1}^{N+2} \bm{z}^n_{cls},
\end{equation}
where $\bm{z}^n_{cls}$ is the $n$th token of $\bm{z}_{cls}$.
Because each task prototype learned from training samples of all of the classes in the task, it contains the global information of all of the classes in the task instead of the information of the specific class in the task and represents discrepancy between tasks.  
Computing the similarity between the pooling feature $\bm{p}_{cls}$ and the task prototypes may not correctly reflect their correlations because each task prototype contains information of multiple classes.
Thus, we map $\bm{p}_{cls}$ by using a fully connected layer which contains $t$ neurons and pass the vector to a softmax function to retrieve the index of the maximal component to indicate the selected task prototype $\bm{\omega}^*_{tsk}$.
%
Similar to the class prompt module, we apply the average task prototype $\bm{\omega}^{avg}_{tsk}$ to maintain the stability of the model as follows:
\begin{equation}
    \bm{\omega}^{avg}_{tsk} = \frac{1}{t} \sum_{p=1}^t \bm{\omega}_p,
\end{equation}
where $\bm{\omega}_p$ is the $p$th task prototype and the dimension of $\bm{\omega}^{avg}_{tsk}$ is $\mathbb{R}^{1 \times D}$.

The selected task prototype $\bm{\omega}^*_{tsk}$ and average task prototype $\bm{\omega}^{avg}_{tsk}$ are concatenated to $\bm{z}_{cls}$ to generate $\bm{z}'_{tsk}$ as follows:
\begin{equation}
    \bm{z}'_{tsk} = \bm{\omega}^*_{tsk} \oplus \bm{\omega}^{avg}_{tsk} \oplus \bm{z}_{cls}.
\end{equation}
where the dimension of $\bm{z}'_{tsk}$ is $\mathbb{R}^{(N+4) \times D}$.

To integrate the task information to class prototypes and patch tokens, we also use the transformer encoder to discover the correlations of these prototypes and tokens. 
%
%
Given $\bm{z}'_{tsk}$, the learned features after the MSA block is denoted as follows:
\begin{equation}
\bm{\hat{z}}_{tsk} = MSA(LN(\bm{z}'_{tsk}))+\bm{z}'_{tsk}.
\end{equation}
$\bm{\hat{z}}_{tsk}$ is then passed to the MLP block to obtain the learned feature $\bm{z}_{tsk}$ with the task prototype information as follows:
\begin{equation}
\bm{z}_{tsk} = MLP(LN(\bm{\hat{z}}_{tsk}))+\bm{\hat{z}}_{tsk}.
\end{equation}

In the first task, the task prototype pool has an initial task prototype with the dimension of $\mathbb{R}^{1 \times D}$ generated by using the Gaussian distribution and this task prototype will be updated during training.   
In task $t$, we create a new task prototype $\bm{\omega}_{t}$ which is the $t$th task prototype and add it to the prototype pool for update.
To avoid catastrophic forgetting of the previous tasks, we freeze task prototypes in the previous task prototype set.

The selected task prototype can be a reminder of the previous task for the transformer encoder when training the images of previous tasks.
Thus, the output features of the task prompt module can better represent the images of the previous tasks and help provide the discrepant features for different tasks. 
Moreover, these task prototypes can also help prevent catastrophic forgetting by considering previously learned task knowledge.
If the task prototype of the new task is selected for the input image, it provides auxiliary information for the new task to improve the adaptability of the model for the new task.
As a result, the proposed method can update the network to learn the information of new tasks, and still remember the information of previous tasks to describe the discrepancy between tasks and prevent the inter-task confusion problem. 

\subsubsection{Unified Classification Module}
In traditional dynamic architecture methods such as~\cite{Yan_CVPR_2021}, new models are directly added to learn images of new classes in each task.
Thus, the number of models will increase with respect to the number of tasks.
More recent methods~\cite{Douillard_CVPR_2022} try to reduce the repetitive learning of models, but still require to learn different classifiers for different tasks.   
%

%
In our method, the task prompt module provides inter-task information.
To simultaneously discriminate all classes by considering the class information, task information and patch tokens, we propose a unified classification module.
In the unified classification module, a patch-wise average pooling layer is used to extract representative features for classification.
The features are passed to a label classifier $c_{label}$ and an auxiliary classifier $c_{aux}$. 
The former classifier aims to classify the input image, while the latter classifier aims to enhance the plasticity of the model for the new task. 
In the following, we will introduce the unified classifier module.

In order to retain the information between prototypes and features, we use patch-wise average pooling to obtain the pooling feature $\bm{p}_{tsk}$ as follows: 
\begin{equation}
    \bm{p}_{tsk} = \frac{1}{N+4} \sum_{n=1}^{N+4} \bm{z}^n_{tsk},
\end{equation}
where $\bm{z}^n_{tsk}$ is the $n$th token of $\bm{z}_{tsk}$. $\bm{p}_{tsk}$ is then passed to the classifier $c_{label}$ to classify all classes from task $1$ to task $t$ as follows: 
\begin{equation}
    \begin{bmatrix}
       \bm{\hat{y}}^{1:t-1}_i \ \bm{\hat{y}}^t_i
    \end{bmatrix}= c_{label}(\bm{p}_{tsk}), 
\end{equation}
where $\bm{\hat{y}}^{1:t-1}_i = (\hat{y}^{1}_i, \hat{y}^{2}_i, \hdots, \hat{y}^{|C^{1:t-1}|}_i)$ represents the prediction probability of classes from task $1$ to task $t-1$, $\bm{\hat{y}}^t_i = (\hat{y}^{|C^{1:t-1}|+1}_i, \hdots, \hat{y}^{|C^{1:t}|}_i)$ represents the prediction probability of the classes in task $t$, and $c_{label}(\cdot)$ represents the classifier of $|C^{1:t}|$ neurons in task $t$ which contains  a fully-connected layer and a softmax function for classification.
In this way, the task information learned in the task prompt module is applied to help the discrepancy between tasks and thus also help reduce inter-task confusion.

In order to overcome the stable-plasticity dilemma, we additionally use an auxiliary classifier $c_{aux}$ to enhance the classification ability of the model for the new task. 
$c_{aux}$ is used to improve the diversity and can enhance the ability to distinguish classes of the current task from the classes of the previous tasks.  
$\bm{p}_{tsk}$ is also passed to the auxiliary classifier to classify $\bm{x}^t_i$ to the current classes in task $t$ or an extra class which indicates $\bm{x}^t_i$ is the input image of the classes of the previous tasks.
The output of $c_{aux}$ is defined as follows: 
\begin{equation}
    \bm{\bar{y}}^t_i = c_{aux}(\bm{p}_{tsk}), 
\end{equation} 
where $\bm{\bar{y}}^t_i = (\bar{y}^{ext}_i, \bar{y}^{|C^{1:t-1}|+1}_i, \hdots, \bar{y}^{|C^{1:t}|}_i)$ is the output of the auxiliary classifier, $\bar{y}^{ext}_i$ is the probability of $\bm{x}^t_i$ that belongs the classes of the previous tasks, the prediction $(\bar{y}^{|C^{1:t-1}|+1}_i, \hdots, \bar{y}^{|C^{1:t}|}_i)$ represents the prediction probability of classes in task $t$, and $c_{aux}(\cdot)$ represents the auxiliary classifier which contains a fully-connected layer and a softmax function for classification.
In this way, $c_{aux}$ gives more ability to learn the new classes to improve the plasticity of the model. 
During inference, we only use $c_{label}$ to obtain the classification results for input images in task $t$. 
As a result, only one classifier is required to classify input images of different classes in previous tasks and task $t$. 

\subsection{Loss Functions}
Three loss functions are applied to drive the learning of the model. 
The first loss is the binary cross entropy loss $\mathcal{L}_{bce}$ used to help the training of $c_{label}$ in task $t$. 
The binary cross entropy loss $\mathcal{L}_i$ for each training sample $\bm{x}^t_i$ is defined as follows:
\begin{equation}
\begin{aligned}
   \mathcal{L}_i = -\frac{1}{|C^{1:t}|}\sum_{j=1}^{|C^{1:t}|}
   {
    {y}^{j}_i \log ( \hat{y}^{j}_i) + (1-{y}^{j}_i) \log (1- \hat{y}^{j}_i),
   } 
\end{aligned}
\end{equation}
where ${y}^{j}_i$ is the ground truth label of $\bm{x}_i^t$ belonging to class $j$.
Then, the binary cross entropy loss $\mathcal{L}_{bce}$ is defined as follows:
\begin{equation}
\mathcal{L}_{bce} = \frac{1}{N^t}\sum_{i=1}^{N^t}\mathcal{L}_i ,
\end{equation}
where $N^t$ is the number of training samples in task $t$.

The second loss is the knowledge distillation loss $\mathcal{L}_{kd}$ to maintain the stability of the model. 
It is calculated based on the outputs of the previous model learned in task $t-1$ and the current model. 
This prevents the model from only focusing on learning new classes in the current task, which will easily cause catastrophic forgetting.
The loss $\mathcal{L}_{kd}$ is defined as follows: 
\begin{equation}
   \mathcal{L}_{kd} = \frac{1}{N^t}\sum_{i=1}^{N^t}KL(\bm{\hat{y}}_i^{1:t-1}, \bm{\tilde{y}}_i^{1:t-1}),
\end{equation}
where $KL(\cdot)$ is the KL divergence function, and $\bm{\hat{y}}^{1:t-1}_i$ and $\bm{\tilde{y}}_i^{1:t-1}$ represent the prediction probability of classes from task $1$ to task $t-1$ of the current model and the previous model, respectively. 

The third loss is the auxiliary loss $\mathcal{L}_{aux}$ which encourages the diversity of learning new classes in task $t$ through the auxiliary classifier. 
$\mathcal{L}_{aux}$ makes the model focus on learning the classes in the current task without forgetting the past knowledge to improve the plasticity of the model.
The auxiliary loss $\mathcal{L}_i^{a}$ for each training sample $\bm{x}^t_i$ and is defined as follows:
\begin{equation}
\begin{aligned}
   \mathcal{L}_{i}^a = -\frac{1}{|C^t|+1}\sum_{k=1}^{|C^t|+1}
    {
    {y}^{k}_i \log ( \bar{y}^{k}_i) + (1-{y}^{k}_i) \log (1- \bar{y}^{k}_i),
   }    
\end{aligned}
\end{equation}
where the ground truth set 
$S =\{{y}^{ext}_i, {y}^{|C^{1:t-1}|+1}_i, \hdots, {y}^{|C^{1:t}|}_i\}$, and 
$y^{ext}_i$ represents if $\bm{x}_i$ belongs to the sample of the previous task. 
Then, the auxiliary loss $\mathcal{L}_{aux}$ is defined as follows:
\begin{equation}
\mathcal{L}_{aux} = \frac{1}{N^t}\sum_{i=1}^{N^t}\mathcal{L}_{i}^a.
\end{equation}
Based on the proposed losses, the model can be effectively trained in an end-to-end manner by considering the previously learned knowledge and new knowledge learned from new classes to overcome the stability-plasticity dilemma. 

\subsection{Implementation Details}
%
%
Our model is optimized by using AdamW~\cite{Loshchilov_ICLR_2019}. 
The batch size is $200$. 
The learning rate is set to $5 \times 10^{-4}$ and the number of epochs is set to $500$. 
The number $L$ of encoders is $11$ and the embedding dimension $D$ is $96$.
The hyperparameter $\lambda$ is $0.1$ for loss computation. 
Our method is implemented by using PyTorch 1.12 and is run with two NVIDIA Tesla V100 GPUs. 
To improve the generalization ability of the learned features, data augmentations~\cite{Chiou_TIP_2023} including random cropping, horizontal flipping, vertical flipping, rotation and MixUp~\cite{Zhang_ICLR_2018} are applied during training.

\section{Experimental Results}
\subsection{Datasets and Settings}
We evaluated our method by using the CIFAR-100~\cite{Krizhevsky_TR_2009}, the ImageNet100 and ImageNet1K datasets~\cite{Deng_CVPR_2009} under different continual learning settings. 
For the CIFAR-100 dataset~\cite{Krizhevsky_TR_2009}, the input image size is $32 \times 32$ and for the ImageNet dataset~\cite{Deng_CVPR_2009}, the input image size is $224 \times 224$. 
The standard continual learning scenarios in CIFAR-100 are 10 steps (10 new classes per task), 20 steps (5 new classes per task) and 50 steps (2 new classes per task). 
In ImageNet100 and ImageNet1K datasets, 10 steps are used for continual learning settings.
Thus, we add 10 new classes per step for the ImageNet100 dataset, and 100 new classes per step for the ImageNet1K dataset.

Besides the top-1 accuracy, we also compare the top-5 accuracy for the ImageNet100 and ImageNet1K datasets. 
The average (Avg) accuracy after each task and the last (Last) accuracy, i.e. the accuracy of the final task, are reported.
A fixed size buffer is used to retain a few samples of the previous tasks $\{1, ..., t-1\}$ which help rehearse the previous samples when training the model in task $t$. 
In the CIFAR-100 and ImageNet100 datasets, we used $2000$ samples in the rehearsal memory and we used $20000$ samples in ImageNet1K.
The aforementioned evaluation settings and metrics for the CIFAR-100, ImageNet100 and ImageNet1K datasets are the same as those in the state-of-the-art methods~\cite{Rebuffi_CVPR_2017, Yan_CVPR_2021, Douillard_CVPR_2022}. 
\subsection{Ablation Study}
\begin{table}[t]
\begin{center}
\caption{The ablation study of prompt modules for CIFAR-100 in 10 steps training setting.}
\begin{tabular}{c c | c c}
  \hline
  \hline
  \thead{Class Prompt \\ Module} & \thead{Task Prompt \\ Module} & Avg & Last \\
  \hline
  \quad & \quad & 71.25 & 54.02 \\
  \checkmark & \quad & 74.60 (+3.35) & 62.11 (+8.09)\\
  \quad & \checkmark & 74.63 (+3.38) & 62.44 (+8.42)\\
  \hline
  \checkmark & \checkmark & \textbf{78.14 (+6.89)} & \textbf{69.57 (+15.55)}\\
  \hline
  \hline
\end{tabular}
\label{tab:ablation}
\end{center}
\end{table}
%
\begin{table}[t]
\begin{center}
\caption{The ablation study of loss functions for CIFAR-100 in 10 steps training setting.}
\begin{tabular}{c c c | c c}
  \hline
  \hline
  $\mathcal{L}_{bce}$ & $\mathcal{L}_{kd}$ & $\mathcal{L}_{aux}$ & Avg & Last \\
  \hline
  \checkmark & \quad & \quad & 76.61 & 62.03 \\
  \checkmark & \checkmark & \quad & 76.60 (-0.01) & 67.37 (+5.34)\\
  \checkmark & \quad & \checkmark & 77.01 (+0.40) & 62.65 (+0.62)\\
  \hline
  \checkmark & \checkmark & \checkmark & \textbf{78.14 (+1.53)} & \textbf{69.57 (+7.54)}\\
  \hline
  \hline
\end{tabular}
\label{tab:ablation2}
\end{center}
\end{table}
%
\begin{table}[t]
\begin{center}
\caption{The ablation study of attentions in the feature encoder module for CIFAR-100 in 10 steps training setting.}
\begin{tabular}{c | c c c c}
  \hline
  \hline
  \thead{Attention Scheme} & Params & Avg & Last \\
  \hline
  MSA & 23.15 & 72.42& 60.64 \\
  \hline
  DiNA & \textbf{10.64} & \textbf{78.14 (+5.72)} & \textbf{69.57 (+8.93)}\\
  \hline
  \hline
\end{tabular}
\label{tab:ablation3}
\end{center}
\end{table}

The ablation study was conducted under the 10 steps scenario in the CIFAR-100 dataset.
We first evaluated the effectiveness of the class prompt module and task prompt module which are the most important components of the proposed method. 
As shown in Table~\ref{tab:ablation}, without the class prompt module and task prompt module, the model is not able to memorize the information of previously learned classes and tasks and thus achieves the worst results especially for the last accuracy. 
With the class prompt module, the learned class prototypes can help the model to memorize the information of previously learned classes and thus avoid the catastrophic forgetting.
In contrast, the task prototypes of the task prompt module provide the discriminate information between tasks to prevent inter-task confusion.
Thus, applying either one of the module helps significantly improve the last accuracy of the model which implies reducing the catastrophic forgetting.
By using both modules, both of the information of learned classes and tasks are reserved during training and provided to boost the classification performance during testing.
As a result, the proposed method achieves the best performance and is shown to solve the catastrophic forgetting problem.

Besides the proposed modules, we also examined the effectiveness of the binary cross entropy loss $\mathcal{L}_{bce}$, the knowledge distillation loss $\mathcal{L}_{kd}$ and the auxiliary loss $\mathcal{L}_{aux}$.
As shown in Table~\ref{tab:ablation2}, the model with only $\mathcal{L}_{bce}$ achieves worse last accuracy compared with the model with $\mathcal{L}_{bce}$ and $\mathcal{L}_{kd}$. 
Such results show that $\mathcal{L}_{kd}$ helps the current model to memorize the information of previously learned classes by making consistent outputs with respect to the outputs of the previous model and thus prevents forgetting previous classes. 

The model with $\mathcal{L}_{bce}$ and $\mathcal{L}_{aux}$ achieves better average accuracy and last accuracy compared with the model with only $\mathcal{L}_{bce}$.
Because $\mathcal{L}_{aux}$ aims to force the model to focus on the classification of new classes, it helps improve the model plasticity to learn classes in new tasks.
In addition, it also considers the information of previous class.
Thus, the last accuracy of the model with $\mathcal{L}_{aux}$ can also be increased. 
By simultaneously applying three losses, the proposed method can achieve the best average accuracy and last accuracy based on the design of $\mathcal{L}_{kd}$ and $\mathcal{L}_{aux}$ in the proposed unified classification module.

In our method, we applied two types of attention schemes in the encoder and the prompt modules. 
In Table~\ref{tab:ablation3}, we compare the impact of using different attention schemes in the feature encoder module.
When we apply the MSA scheme instead of the DiNA scheme in the feature encoder module, the number of parameters will significantly increase. 
Moreover, the performance of using the MSA scheme is undermined, which proves that applying MSA in the feature encoder for the continual learning task will face overfitting problems as shown in~\cite{Touvron_2021_ICML}.
Thus, the DiNA scheme is suggested to extract features in the feature encoder module.

\subsection{Quantitative Results}
\begin{table*}[t]
  \caption{The results of the competing methods and the proposed method for CIFAR-100 in three scenarios.}
  \begin{center}
    \begin{tabular}{c | c | c c c | c c c | c c c }
      \hline
      \hline
      & & \multicolumn{3}{c|}{10 steps} & \multicolumn{3}{c|}{20 steps} & \multicolumn{3}{c}{50 steps}\\
      Methods & Category & Params & Avg & Last & Params & Avg & Last & Params & Avg & Last\\
      \hline
      ResNet18 Joint~\cite{Douillard_CVPR_2022} & - & 11.22 & - & 80.41 & 11.22 & - & 81.49 & 11.22 & - & 81.74\\
      Transformer Joint~\cite{Douillard_CVPR_2022} & - & 10.72 & - & 76.12 & 10.72 & - & 76.12 & 10.72 & - & 76.12\\
      \hline
      iCaRL~\cite{Rebuffi_CVPR_2017} & rehearsal & 11.22 & 65.27 & 50.74 & 11.22 & 61.20 & 43.75 & 11.22 & 56.08 & 36.62 \\
      UCIR~\cite{Hou_CVPR_2019} & regularization & 11.22 & 58.66 & 43.39 & 11.22 & 58.17 & 40.63 & 11.22 & 56.86 & 37.09 \\
      BiC~\cite{Wu_CVPR_2019} & regularization & 11.22 & 68.80 & 53.54 & 11.22 & 66.48 & 47.02 & 11.22 & 62.09 & 41.04 \\
      WA~\cite{Zhao_CVPR_2020} & regularization & 11.22 & 69.46 & 53.78 & 11.22 & 67.33 & 47.31 & 11.22 & 64.32 & 42.14 \\
      PODNet~\cite{Douillard_ECCV_2020} & regularization & 11.22 & 58.03 & 41.05 & 11.22 & 53.97 & 35.02 & 11.22 & 51.19 & 32.99 \\
      RPSNet~\cite{Rajasegaran_NIPS_2019} & dynamic & 56.50 & 68.60 & 57.05 & - & - & - & - & - & - \\
      DER w/o P~\cite{Yan_CVPR_2021} & dynamic & 112.27 & 75.36 & 65.22 & 224.55 & 74.09 & 62.48 & 561.39 & 72.41 & 59.08 \\
      DyTox+~\cite{Douillard_CVPR_2022} & dynamic & 10.73 & 75.54 & 62.06 & 10.74 & 75.04 & 60.03 & 10.77 & 74.35 & 57.09 \\
      TSM~\cite{Huang_MVA_2023} & dynamic & 10.73 & 74.30 &  66.38 & 10.73 & 71.89 & 61.40 & 10.75 & 70.30 & 56.72 \\
      \hline
      Proposed & dynamic & 10.64 & \textbf{78.14} & \textbf{69.57} & 10.64 & \textbf{76.34} & \textbf{65.62} & 10.65 & \textbf{74.68}&  \textbf{61.14} \\
      \hline
      \hline
    \end{tabular}
  \end{center}
  \label{tab:CIFAR}
\end{table*}
\begin{table*}[t]
  \caption{The results of the competing methods and the proposed method for ImageNet100 and ImageNet1K.}
  \begin{center}
    \begin{tabular}{c | c | c c c c c | c c c c c }
      \hline
      \hline
      & & \multicolumn{5}{c|}{ImageNet100 10 steps} & \multicolumn{5}{c}{ImageNet1K 10 steps}\\
      & & & \multicolumn{2}{c}{top-1} & \multicolumn{2}{c|}{top-5} & & \multicolumn{2}{c}{top-1} & \multicolumn{2}{c}{top-5}\\  
      \cline{4-7} \cline{9-12}
      Methods & Category & Params & Avg & Last & Avg & Last & Params & Avg & Last & Avg & Last\\
      \hline
      ResNet18 Joint~\cite{Douillard_CVPR_2022} & - & 11.22 & - & - & - & 95.10 & 11.68 & - & - & - & 89.27 \\
      Transformer Joint~\cite{Douillard_CVPR_2022} & - & 11.00 & - & 79.12 & - & 93.48 & 11.35 & - & 73.58 & - & 90.60 \\
      \hline
      iCaRL~\cite{Rebuffi_CVPR_2017} & rehearsal & 11.22 & - & - & 83.60 & 63.80 & 11.68 & 38.40 & 22.70 & 63.70 & 44.00 \\
      E2E~\cite{Castro_ECCV_2018} & rehearsal & 11.22 & - & - & 89.92 & 80.29 & 11.68 & - & - & 72.09 & 52.29 \\
      BiC~\cite{Wu_CVPR_2019} & regularization & 11.22 & - & - & 90.60 & 84.40 & 11.68 & - & - & 84.00 & 73.20 \\
      WA~\cite{Zhao_CVPR_2020} & regularization & 11.22 & - & - & 91.00 & 84.10 & 11.68 & 65.67 & 55.60 & 86.60 & 81.10 \\
      RPSNet~\cite{Rajasegaran_NIPS_2019} & dynamic & - & - & - & 87.90 & 74.00 & - & - & - & - & - \\
      DER w/o P~\cite{Yan_CVPR_2021} & dynamic & 112.27 & 77.18 & 66.70 & 93.23 & 87.52 & 116.89 & 68.84 & 60.16 & 88.17 & 82.86 \\
      Simple-DER~\cite{Li_arXiv_2021} & dynamic & - & - & - & - & - & 28.00 & 66.63 & 59.24 & 85.62 & 80.76 \\
      DyTox+~\cite{Douillard_CVPR_2022} & dynamic & 11.01 & 77.15 & 69.10 & 92.04 & 87.98 & 11.36 & 71.29 & 63.34 & 88.59 & 84.49 \\
      \hline
      Proposed & dynamic & 10.64 & \textbf{81.54} & \textbf{72.48} & \textbf{94.73} & \textbf{91.56} & 11.05 & \textbf{76.13} & \textbf{66.08} & \textbf{92.39} & \textbf{88.19} \\
      \hline
      \hline
    \end{tabular}
  \end{center}
  \label{tab:ImageNet}
\end{table*}

%
We compared the proposed method with state-of-the-art continual learning methods in the CIFAR-100, ImageNet100 and ImageNet1K datasets, respectively.
The state-of-the-art rehearsal methods including  iCaRL~\cite{Rebuffi_CVPR_2017} and E2E~\cite{Castro_ECCV_2018}, regularization methods including UCIR~\cite{Hou_CVPR_2019}, BiC~\cite{Wu_CVPR_2019}, WA~\cite{Zhao_CVPR_2020}, and PODNet~\cite{Douillard_ECCV_2020}, and dynamic architecture methods including  RPSNet~\cite{Rajasegaran_NIPS_2019}, DER without pruning (w/o P)~\cite{Yan_CVPR_2021}, Simple-DER~\cite{Li_arXiv_2021}, DyTox+~\cite{Douillard_CVPR_2022} and TSM~\cite{Huang_MVA_2023} are compared.

Table~\ref{tab:CIFAR} shows the results of the competing methods and the proposed method for the CIFAR-100 dataset.
The ResNet18 Joint and Transformer Joint~\cite{Douillard_CVPR_2022} learn the training data of all classes and tasks by using ResNet18 and vision transformer~\cite{Dosovitskiy_ICLR_2021} under similar numbers of parameters, respectively.
Thus, the catastrophic forgetting problem can be avoided and the results of these two networks serve as the upper bound of model performance in the continual learning settings.
Because the rehearsal method, 
iCaRL~\cite{Rebuffi_CVPR_2017}, relies on the stored exemplars of learned classes to avoid catastrophic forgetting, its performance is limited by the memory sizes.
To avoid storing too many exemplars, the regularization methods~\cite{Hou_CVPR_2019,Wu_CVPR_2019,Zhao_CVPR_2020,Douillard_ECCV_2020} are proposed.
They limit the model update to avoid forgetting information of previous tasks.
Thus, when the number of steps increases, the last accuracy of the regularization methods such as BiC~\cite{Wu_CVPR_2019} and WA~\cite{Zhao_CVPR_2020} significantly drops, which implies these methods fall into the stability-plasticity dilemma. 
As a result, the performance of the rehearsal and regularization methods are not satisfactory.

To better learn each task, the dynamic architecture methods~\cite{Rajasegaran_NIPS_2019,Yan_CVPR_2021,Douillard_CVPR_2022,Huang_MVA_2023} learn each task by using specific designed models.
With the increasing number of tasks, the number of models will also increase to learn new tasks as shown in RPSNet~\cite{Rajasegaran_NIPS_2019} and DER w/o P~\cite{Yan_CVPR_2021}.
Thus, these dynamic architecture methods can achieve better model plasticity when learning new tasks and outperform the rehearsal and the regularization methods.
However, conventional dynamic architecture methods such as RPSNet~\cite{Rajasegaran_NIPS_2019} and DER w/o P~\cite{Yan_CVPR_2021} require a large number of parameters to learn new tasks, when the number of steps increases. 
To solve the parameter problem, DyTox+~\cite{Douillard_CVPR_2022} reuses the learned model to learn new tasks and trains new classifiers for each task. 
During inference, all of the learned classifiers are used to identify the label of the input image.
TSM~\cite{Huang_MVA_2023} aims to reuse the learned model and learned classifier to solve the parameter problem by using task selection module.
However, TSM may fail to select incorrect tasks during inference, and thus achieves worse results in average accuracy compared with DyTox+~\cite{Douillard_CVPR_2022}.

Our method solves the catastrophic forgetting by using the class prompt module and task prompt module.
The class prompt module retains information of previous classes to avoid catastrophic forgetting while also giving our model the plasticity to learn new classes prototypes of new classes. 
To distinguish different classes from different tasks, the task prompt module further learns the task information by considering the task prototypes. 
By simultaneously maintaining the prototypes of previous tasks and learning prototypes of new tasks, our method can better represent each task and the classes in each task. 
As a result, only a single encoder module and a unified classification module are required to solve the continual learning problem in our method, which leads to better average and last accuracy and a nearly fixed number of model parameters as shown in Table~\ref{tab:CIFAR}. 

As shown in Table~\ref{tab:ImageNet}, most dynamic architecture methods~\cite{Yan_CVPR_2021,Li_arXiv_2021,Douillard_CVPR_2022} are still more effective in solving the catastrophic forgetting problem compared with the rehearsal methods~\cite{Rebuffi_CVPR_2017,Castro_ECCV_2018} and the regularization methods~\cite{Wu_CVPR_2019,Zhao_CVPR_2020} for the ImageNet100 dataset.
Compared with the competing dynamic architecture methods, our method achieves the best top-1 and top-5 average and last accuracy, respectively. 
When the number of classes increases in each task in the ImageNet1K dataset, the average and last accuracy of the methods in the ImageNet1K dataset drop compared with those in the ImageNet100 dataset. 
Nevertheless, the proposed method still achieves significant better top-1 and top-5 average and last accuracy.
With the ImageNet100 and ImageNet1K datasets, we show the effectiveness of the proposed class prompt module and task prompt module to overcome the stability-plasticity dilemma and solve the catastrophic forgetting problem in continual learning with few parameters.

\begin{figure}[!t]
\centering
\includegraphics[width=0.48\textwidth]{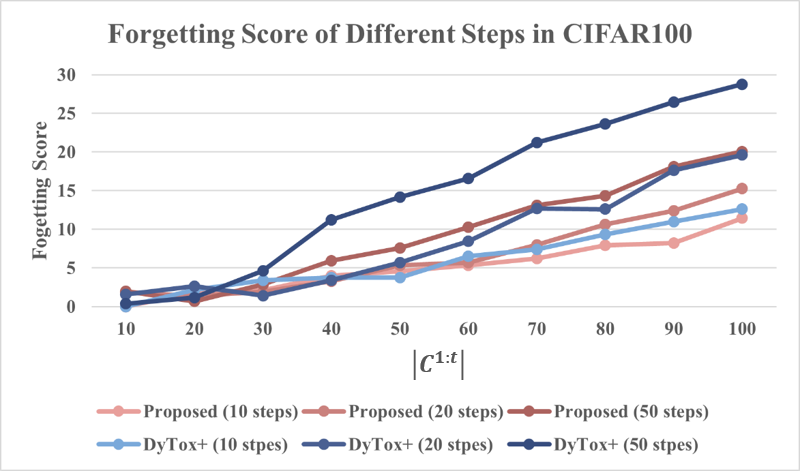}
\caption{Comparison of Forgetting Scores in the CIFAR-100 dataset with different steps.
}
\label{fig:forget}
\end{figure}

\subsection{Discussions}
To demonstrate the superiority of the proposed method in solving the catastrophic forgetting problem, we compared the forgetting scores~\cite{Chaudhry_2018_ECCV} of the proposed method with the second best method DyTox+~\cite{Douillard_CVPR_2022}.
We first define the forgetting score for a learned class as the difference between the best testing accuracy of the class by using the previous model throughout the continual learning process and the testing accuracy of the class by using the current model in task $t$. 
Thus, the forgetting score gives an estimation of how the current model forgets about the class compared with the best previous model. 
We define the forgetting score $f_j^t$ of the $j$th class in task $t$ as follows:
\begin{equation}
    f_j^t = a_j^{max} - a_j^t, \quad \forall j \leq |C^{1:t-1}|,
\end{equation}
where
$a_j^{max}$ is the best testing accuracy of class $j$ by using previous models, $a_j^t$ is the testing accuracy of class $j$ by using the current model learned in task $t$, and $|C^{1:t-1}|$ is the number of previously learned classes before task $t$.
Please note that $f_j^t \in \begin{bmatrix} -1, 1 \end{bmatrix}$ is defined for $j \leq |C^{1:t-1}|$ as we are interested in quantifying forgetting scores for previous tasks.
By normalizing against the number of previous classes, the average forgetting score in task $t$ is defined as follows:
\begin{equation}
    f^t_{class} = \frac{1}{|C^{1:t-1}|} \sum_{j=1}^{|C^{1:t-1}|}{f^t_j}.
\end{equation}
A larger forgetting score $f^t_{class}$ indicates that the model has more serious catastrophic forgetting problem.

As shown in Fig.~\ref{fig:forget}, we compare the forgetting scores with DyTox+ in the CIFAR-100 dataset.
The experiments are performed under three training settings with a total of 10, 20 and 50 steps, respectively.
In general, a training setting with more steps will make the model easily forget previous classes and result in higher forgetting scores.
With the previous knowledge learned by class prototypes and task prototypes, the proposed method has slightly lower forgetting scores under the 10 steps scenario compared with DyTox+.
When the numbers of steps increase to 20 and 50, the forgetting scores of both methods increase to reflect the fact that more classes and tasks are harder to be learned. 
However, the forgetting scores of the proposed method under the 50 steps scenario are even equivalent to those of DyTox+ under the 20 steps scenario.
Compared with DyTox+, the proposed method has significant lower forgetting scores because of the proposed class prompt module and task prompt module.
This figure shows promising results of the proposed method to effectively solve the catastrophic forgetting problem with both modules to retain learned class and task knowledge.

\section{Conclusions}
In this paper, we propose a dynamic prompt transformer (DPFormer) to solve the catastrophic forgetting problem and the stability-plasticity dilemma in continual learning.
To avoid catastrophic forgetting, the class prompt module helps effectively remember learned class information, while the task prompt module helps 
describe the discrepancy between tasks and prevent inter-task confusion.
Moreover, both modules simultaneously learn to represent new classes and tasks by using class prototypes and task prototypes with attention schemes in transformer encoders. 
Our method can effectively preserve past knowledge and give the model the ability to continuously learn new knowledge, making our method more suitable for learning with a large number of tasks.
Therefore, it also overcomes the stability-plasticity dilemma while preventing catastrophic forgetting.
In the future, we aim to extend the proposed method to solve the semantic segmentation problem in continual learning. 

\section*{Acknowledgments}
The authors would like to thank National Center for High-performance Computing (NCHC) for providing computational and storage resources.

\bibliographystyle{IEEEtran}
\bibliography{2023_DPFormer}


\begin{IEEEbiography}[{\includegraphics[width=1in,height=1.25in,clip,keepaspectratio]{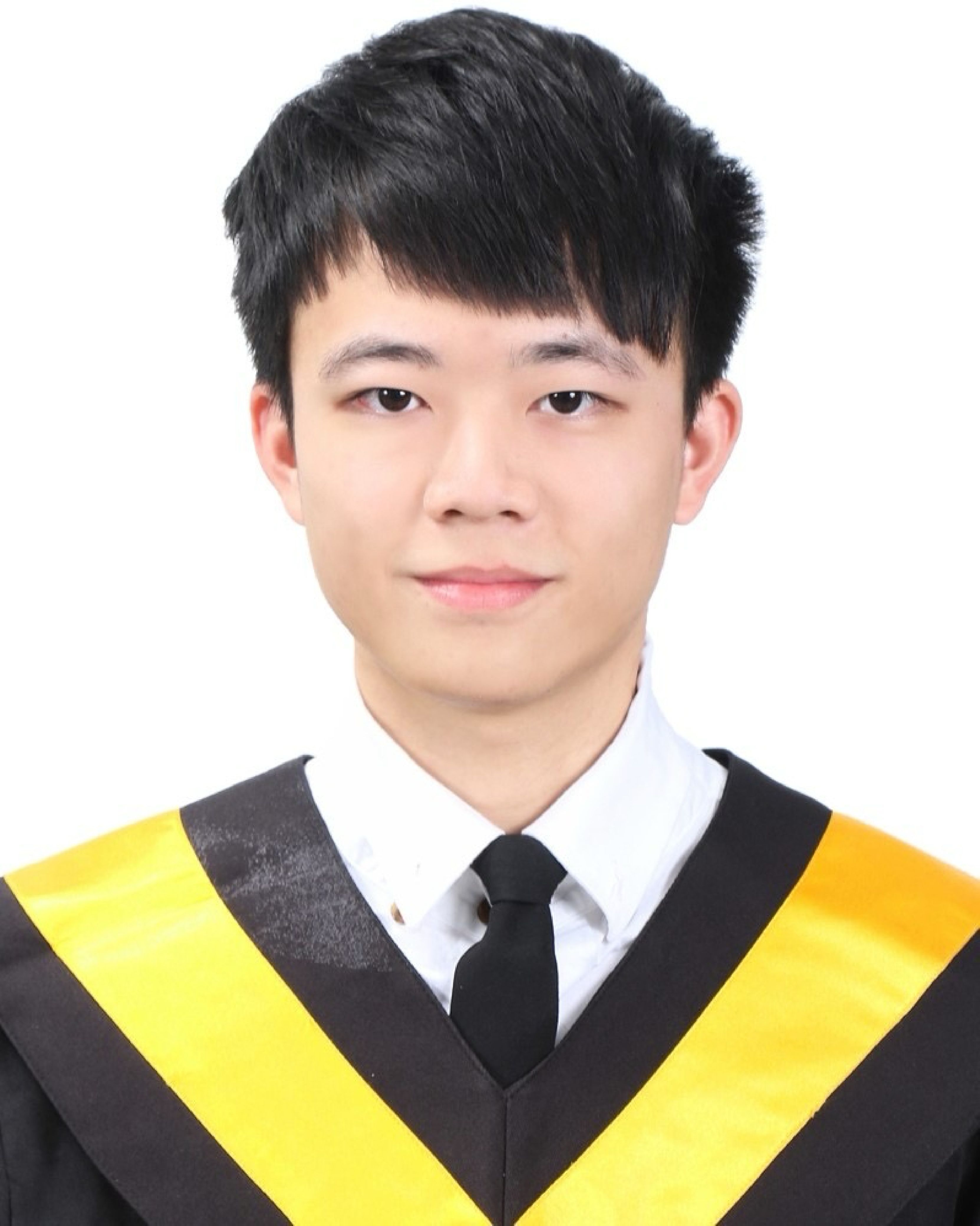}}]{Sheng-Kai Huang} received the B.S. degrees from the Department of Computer Science and Information Engineering, National University of Kaohsiung, Kaohsiung, Taiwan, in 2021 and M.S. degrees from the Department of Computer Science and Engineering, National Chung Hsing University, Taichung, Taiwan, in 2023. He joined with the Department of Software-GPU, Nvidia Corporation, Taiwan in 2023. \end{IEEEbiography}

\begin{IEEEbiography}[{\includegraphics[width=1in,height=1.25in,clip,keepaspectratio]{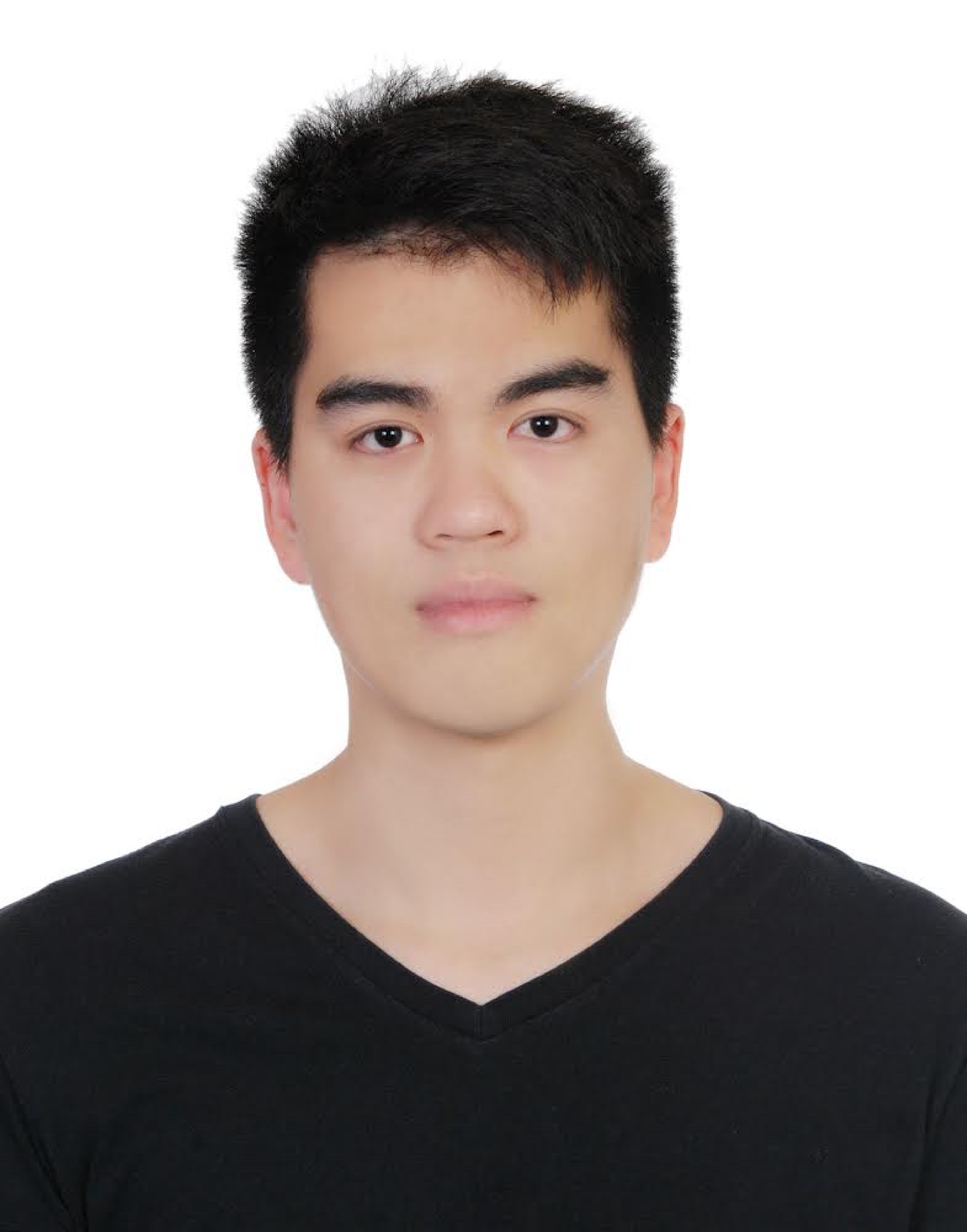}}]{Jiun-Feng Chang} received the B.S. degree from the Department of Computer Science and Engineering, National Chung Hsing University, Taichung, Taiwan, in 2023. Since 2023, he studied for the M.S. degrees in the Department of Computer Science and Engineering, National Chung Hsing University, Taichung, Taiwan. 
\end{IEEEbiography}

\begin{IEEEbiography}[{\includegraphics[width=1in,height=1.25in,clip,keepaspectratio]{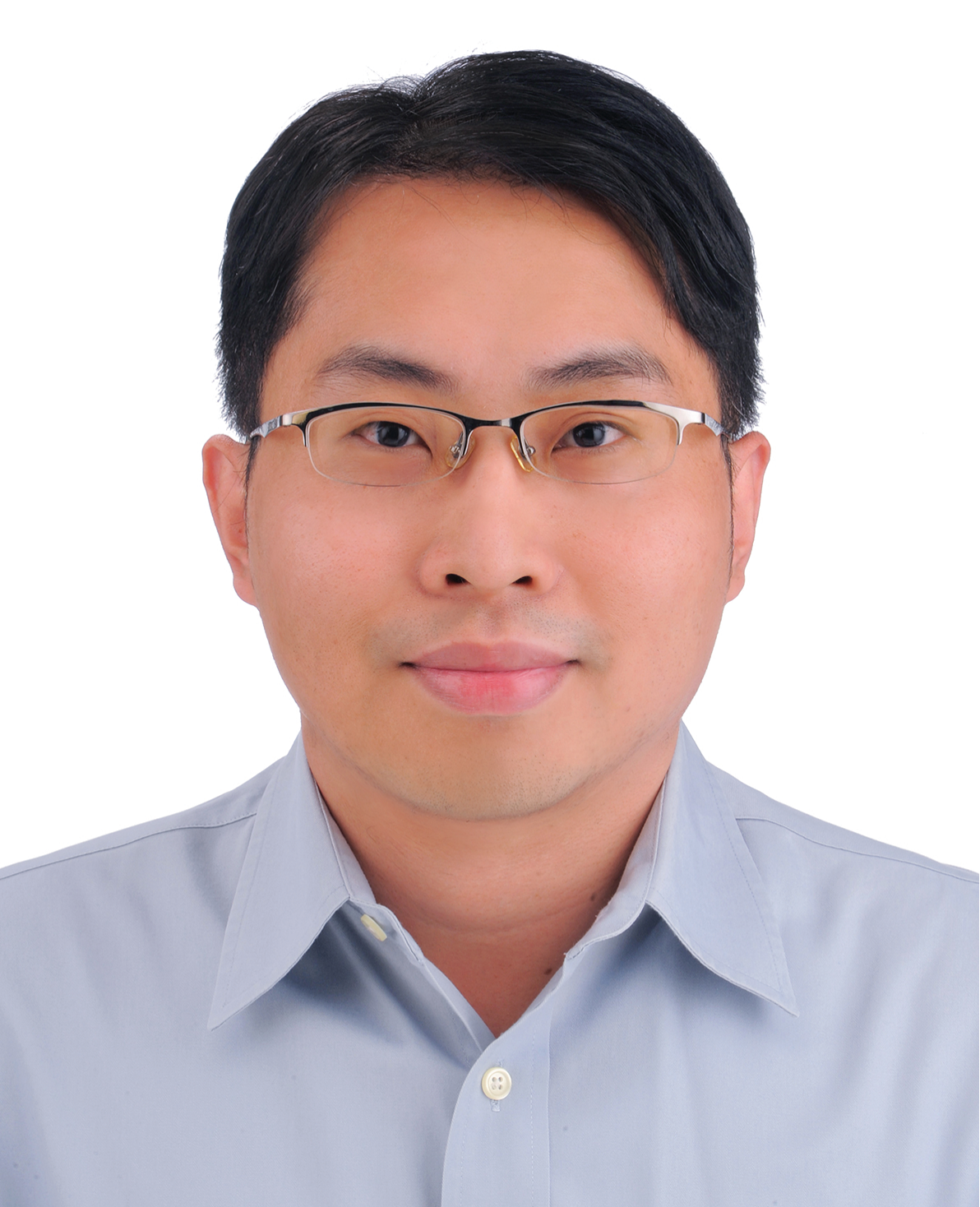}}]{Chun-Rong Huang} (Senior Member, IEEE) received the B.S. and Ph.D. degrees from the Department of Electrical Engineering, National Cheng Kung University, Tainan, Taiwan, in 1999 and 2005, respectively. In 2005, he joined the Institute of Information Science, Academia Sinica, Taipei, Taiwan, as a Postdoctoral Fellow. He joined the Department of Computer Science and Engineering, National Chung Hsing University, Taichung, Taiwan, in 2010, where he became a Full Professor in 2019, and the Cross College Elite Program, and Academy of Innovative Semiconductor and Sustainable Manufacturing, National Cheng Kung University, Tainan, Taiwan in 2023. In 2024, he joined the Department of Computer Science, National Yang Ming Chiao Tung University, Hsinchu, Taiwan.
His research interests include computer vision, computer graphics, multimedia signal processing, image processing, and medical image processing. He is a member of the IEEE Circuits and Systems Society, the IEEE Signal Processing Society, the IEEE Computational Intelligence Society, and the Phi Tau Phi Honor Society.
\end{IEEEbiography}

\end{document}